\documentclass{bmvc2k}
\usepackage{multirow}
\usepackage{amsmath,amssymb} 
\usepackage{floatrow}
\usepackage{lipsum}
\usepackage{graphicx}
\newfloatcommand{capbtabbox}{table}[][\FBwidth]

\title{CTRN: Class Temporal Relational Network for Action Detection}

\addauthor{Rui Dai}{rui.dai@inria.fr}{1}
\addauthor{Srijan Das}{srijan.das@stonybrook.edu}{2}
\addauthor{François Brémond}{francois.bremond@inria.fr}{1}

\addinstitution{
Inria, Université Côte d'Azur,\\
France
}
\addinstitution{
Stony Brook University,\\USA
}
\runninghead{Dai, Das, Bremond}{CTRN}

\begin{document}

\maketitle

\begin{abstract}
Action detection is an essential and challenging task, especially for densely labelled datasets of untrimmed videos. 
There are many real-world challenges in those datasets, such as composite action, co-occurring action, and high temporal variation of instance duration. 
For handling these challenges, we propose to explore both the class and temporal relations of detected actions. 
In this work, we introduce an end-to-end network: Class-Temporal Relational Network (CTRN). 
It contains three key components: (1) The Representation Transform Module filters the class-specific features from the mixed representations to build a graph structured data. (2) The Class-Temporal Module models the class and temporal relations in a sequential manner. (3) G-classifier leverages the privileged knowledge of the snippet-wise co-occurring action pairs to further improve the co-occurring action detection. 
We evaluate CTRN on three challenging densely labelled datasets and achieve state-of-the-art performance, reflecting the effectiveness and robustness of our method. 
\end{abstract}

\section{Introduction}
\label{sec:intro}
\indent Action detection is a challenging computer vision problem which targets at finding precise temporal boundaries of actions occurring in an untrimmed video.
Many studies on action detection focus on videos with sparse and well-separated instances of action~\cite{gtad,PGCN2019ICCV,AFNET}. For instance, action detection algorithms on popular datasets like THUMOS~\cite{THUMOS14} and ActivityNet~\cite{caba2015activitynet} generally learn representations for single actions in a video.
However, in daily life, human actions are continuous and can be very dense. Every minute is filled with potential actions to be detected and labelled. The methods designed for sparsely labelled datasets are hard to generalize to such real-world scenarios. 

\begin{figure*}[t]
\centering
\includegraphics[width=0.8\linewidth]{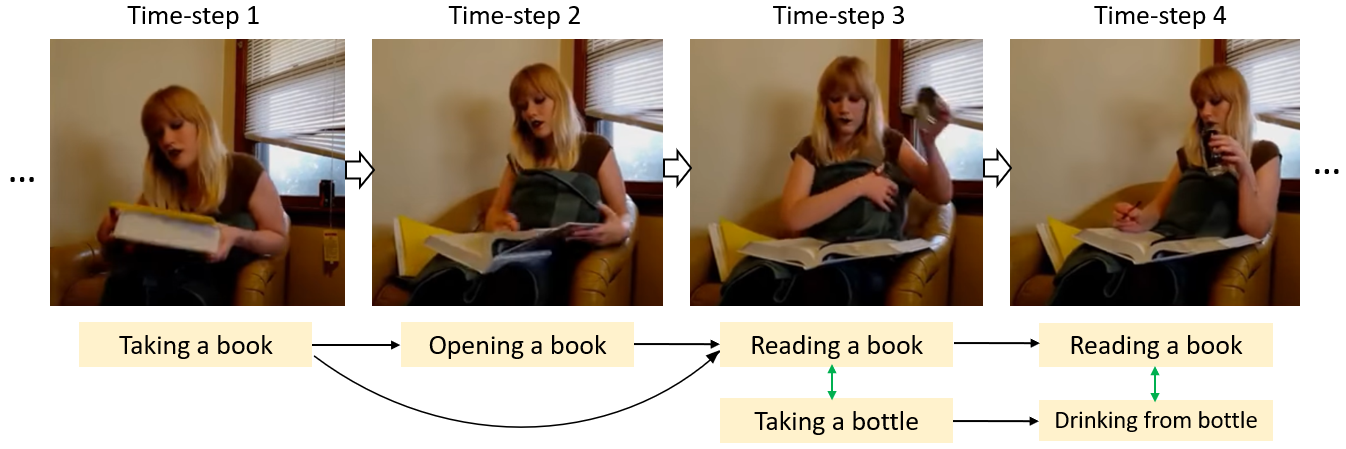}
\caption{Class-Temporal Relation. In a dense labelled video, there are dependencies between action classes (1) across \textbf{different} time steps in {\color{black}black} arrows and (2) at \textbf{same} time step (i.e. co-occurring actions) in {\color{green}green} arrows.}
\label{Fig:front}
\end{figure*}

Towards this research direction, several methods~\cite{TGM1,Dai_2021_WACV,superevent} have been proposed to model complex temporal relationships and to process datasets like Charades~\cite{charades}, TSU~\cite{dai2020toyota} and MultiTHUMOS~\cite{multi-thumos}.
Those datasets encompassing real-world challenges share the following characteristics:
Firstly, the actions are densely labelled and background instances are rare in these videos compared to sparsely labelled datasets. 
Secondly, the video has rich temporal structure and a set of actions occurring together often follows a well defined temporal pattern. For example, \textit{drinking from bottle} always happens after \textit{taking a bottle} and \textit{reading a book} also related to \textit{opening a book} in Fig~\ref{Fig:front}.
Finally, humans are great at multitasking, multiple actions can co-occur at the same time. For example, \textit{reading book} while \textit{drinking water}.

Existing methods have mostly focused on modelling the variation of visual cues across time locally~\cite{lea2017temporal} or globally~\cite{superevent} within a video. However, these methods take into account the temporal information without any further semantics. Real-world videos contain many complex actions with inherent relationships between action classes at the same time steps or across distant time steps (see Fig.~\ref{Fig:front}). Modelling such class-temporal relationships can be extremely useful for locating actions in those videos. 
%
%
%
%

%

To this end, we introduce Class-Temporal Relational Network (CTRN) to harness the relationships among the action classes in a video. 
To explore such relations, CTRN first filters the class-specific representation from the input features at each time step in a video. 
Then the transformed per-class representation is utilized for modelling the inter-class relations. 
(1) Across \textbf{different} time-steps, a graph-based layer is proposed to learn the dependencies between different action classes of the video. 
This learned relation map is shared among all the time-steps to refine the action features from the related actions (e.g. \textit{open the book} and \textit{read book}). Then a temporal layer is used to aggregate features from the same class over time to enable the graph-based layer to explore both short-term and long-term class dependencies. 
(2) At the \textbf{same} time step, a graph-based classifier is proposed to leverage the privileged co-occurring action probabilities to improve co-occurring action detection.  


To summarize, the main contributions of this work are:
1) A graph-based module to explore action class relations across different time-steps.
2) A graph-based classifier to tackle the co-occurring action challenge. 
3) We evaluate our model on three challenging densely labelled datasets for the action detection task. Our method outperforms state-of-the-art results using fewer parameters and FLOPs. 

\vspace{-0.2in}

\section{Related Work}
Action detection has received a lot of interest in recent years~\cite{Dai_2021_ICCV,huang2020improving,dai2019self,superevent}. 
In this work, we focus on densely labelled action detection for handling videos with additional temporal relationships between different action classes~\cite{multi-thumos,charades,dai2020toyota}. 
Different from the sparsely labelled detection methods which output a sparse set of action snippets~\cite{caba2015activitynet,THUMOS14}, densely labelled detection methods need to predict what action is occurring at every snippet in a video. 
There are two principle frameworks for densely labelled action detection: the anchor-based~\cite{RC3d,AFNET} and the Seq2Seq-based~\cite{lea2017temporal,dai2019self} frameworks. 
The anchor-based frameworks are often slow and suffer from
over-generated proposals and rigid boundaries. 
Seq2Seq-based~\cite{TGM1} frameworks apply temporal filters over snippet-wise features with a snippet-level classification, therefore, it interprets the sequence of images to a sequence of predictions. 
Compared to anchor-based methods, the Seq2Seq methods achieve better performance for action detection on datasets with densely distributed actions. This is mainly because of the combinatorial explosion of action proposals generated by the former methods. Thus, the recent methods are following the Seq2Seq framework:
Lea et al.~\cite{lea2017temporal} introduced temporal convolutional network (TCN) for the action detection task. This method increases the temporal reception field by using dilated convolutions to model long temporal patterns.
Similarly, Piergiovanni et al.~\cite{TGM1} introduced Temporal Gaussian Mixture (TGM) layers. In contrast to standard convolution layer, TGM computes the filter weights based on Gaussian distributions, which enables TGM to learn longer temporal structures with a limited number of parameters. 
However, both methods focus only on the temporal modelling while overlooking the action class relations in the untrimmed videos. 
Although Super-event~\cite{superevent} models the 
latent contextual representation among actions, the modelled Super-event represents only the latent correlation among the time steps and not among the action classes. 
%
Most related to our research direction, Tirupattur et al.~\cite{MLAD} introduced MLAD that can explore the class-temporal relations with a set of self-attention layers: an inter-class attention map for every time-step and an inter-time attention map for every action class.   
However, the large number of attention maps leads to huge computational costs for long untrimmed videos and hence, limits the model to learn the discriminative relations among the action classes.
%
%
%
%
{To tackle this, we propose CTRN, which is a graph-based model. 
%
%
Different from MLAD, CTRN explores the action class relation shared by all the time steps but in different temporal scales. This design enables CTRN to effectively handle both short-term and long-term action relations simultaneously.  
Moreover, we propose a G-Classifier that focuses on the co-occurring actions taking into account the inter-dependencies of the actions in the training distribution.
The proposed method is introduced in details in the following section. 
}

%

\section{Proposed Method}
In this section, we formulate the proposed end-to-end model Class-Temporal Relational Network (CTRN) for action detection. 
As shown in Fig.~\ref{Fig:overall}, our model is composed of four major components.
The \textbf{Visual Encoder} encodes the video into a sequence of snippet-level spatio-temporal representation. 
This representation is fed to a Class-temporal Relational Network (CTRN) that predicts the action labels at each time instant. The sub-components in CTRN consist of the following:
Firstly, a \textbf{Representation Transform Module}, which transforms the mixed visual representation into a class-wise representation. 
Secondly, a \textbf{Class-Temporal Module} explores the action class relations across different time-steps and at different temporal resolutions. 
%
%
Finally, a \textbf{G-Classifier} which classifies the class-temporal features into action categories. Unlike previous binary classifiers~\cite{superevent,TGM1} that overlook the dependencies between the action classes, G-Classifier leverages the privilege class dependencies within the training data, thus improving the co-occurring action detection performance. 
In the following, we introduce these modules in details. 
\subsection{Visual Encoder}
Similar to most action detection models~\cite{lea2017temporal,superevent,TGM1}, our model processes the features on top of video snippet representations extracted from 2D/3D CNNs. 
In this work, we use spatio-temporal features extracted from RGB and Optical Flow (OF) I3D networks~\cite{i3d} to encode appearance and motion information respectively. Then, a video is divided into $T$ non-overlapping snippets, each snippet consisting of 16 frames. The inputs to the RGB and Flow deep networks are either the color images or the corresponding OF frames of a snippet. 
We stack the snippet-level features along the temporal axis to form a $T\times D_1$ dimensional video representation, denoted as $X$. 
The action instances in $X$ are always longer than a snippet and their visual representation mixes information of all action classes. As a result, $X$ is not discriminative enough, and needs both temporal and class modelling. {To this end, we develop the class-temporal relationship from the input representation $X$ within CTRN which is described in the following.
Note that the model architecture remains the same for both RGB and OF streams.} \looseness=-5

\subsection{Representation Transform Module} 
\begin{figure*}[t]
\centering
\includegraphics[width=\linewidth]{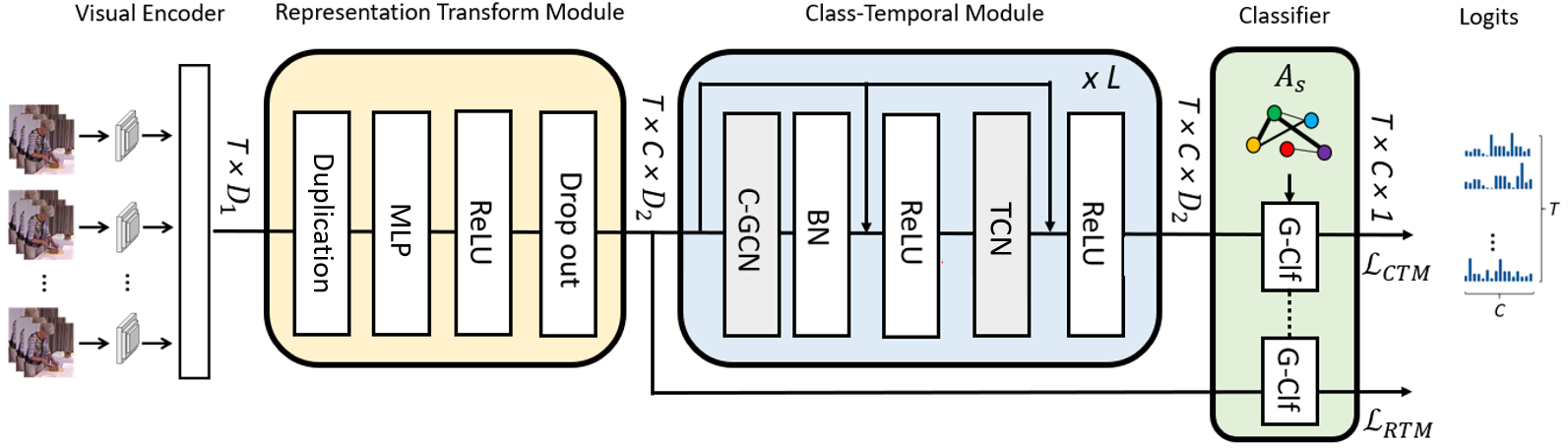}
\caption{Overall structure. The model composed of a Visual 
Encoder, a Representation Transform Module, a Class-Temporal Module (with C-GCN and TCN) and a G-Classifier (i.e. G-Clf). Note: Two G-Clfs are sharing the weights. }
\label{Fig:overall}
\end{figure*}

The input $X$ is first fed into the Representation Transform Module (RTM). The goal of this module is to transform the input to a class-specific representation and to lightweight the channel size to facilitate the following computation.
In practice, RTM duplicates the input features $C$ times into a new dimension representing the action classes followed by a channel-mixer MLP with non-linear activation and dropout.MLP is the linear transformation layer~\cite{almeida1997c1} to do the linear projection.
The equation can be formulated as:
\begin{equation}
\small
    X'_i=ReLU(MLP(X))
\end{equation}
\vspace{-2mm}
\begin{equation}
\small
    X' = DropOut([X'_1, X'_2,...X'_C]))
\end{equation}
where $X'\in \mathbb{R}^{T \times C \times D_2}$ is the output representation of RTM. $D_2 = \frac{D_1}{\beta}$ in which $\beta$ is larger than 1 to shallow the channel size.  
 In order to learn class-specific representation, we embed an auxiliary branch with a G-classifier that maps $X'$ to the action labels (see Fig.~\ref{Fig:overall}).
This transformed feature representation is further exploited to explore the class and temporal relations in the subsequent modules of the network. 

\subsection{Class-Temporal Modeling}
The Class-Temporal Module (CTM) is the key component of CTRN that exploits the class-temporal relations of its input feature. Inspired by the recent success of Graph Convolutional Network (GCN) in relational reasoning~\cite{kipf2016semi,PGCN2019ICCV,gtad,huang2020improving}, we build this module with GCN.  
The objective of this component is to update the feature representations by propagating the information across different classes and across different time steps. 
For modelling the action class relations, we introduce a Class-GCN (C-GCN) layer while the traditional Temporal Convolutional Network (TCN) layer~\cite{lea2017temporal} is utilized to aggregate the temporal information. 
The combination of C-GCN and TCN enables CTM to capture the class semantic information along different temporal hierarchies. Thanks to the learnable graph structure, C-GCN is adaptive with the temporal scale set by TCN. 
%
%
%
%
%

%
In the following, we first introduce how we map the feature representation to the graph structure and then we introduce the CTM components.

\subsubsection{Representation-to-Graph Mapping} 
\label{sec:R2G}
For GCN to process the action relations, the data is to be converted into a graphical structure.
As we have transformed the representation into the class-specific format, thus each vertex of the graph represents an action class at a time step with an embedding vector belonging to $ \mathbb{R}^{D_2}$. 
In total, the graph consists of $C \times T$ vertices whose topology is defined by an adjacency matrix ($A_\mathcal{C}$). This matrix determines whether there are connections (i.e. relations) and its weights determine the intensity of the connections.
%
%
%

\subsubsection{Class-GCN (C-GCN)}
Class-GCN aims at performing the cross-class reasoning over the constructed graph representation. 
The relations between the many action instances are complex and are different across videos. Besides, multiple C-GCNs are stacked in CTM through which C-GCNs capture different levels of semantic information. 
Consequently, the graph adjacency $A_\mathcal{C}$ learns from the data itself for it to be adaptive across different temporal scales. 

In practice, $A_\mathcal{C} \in \mathbb{R}^{C\times C}$ is parameterized and is optimized together with other parameters in the training process. 
Moreover, to differentiate the class relations owing to different videos, the adjacency matrix $A_\mathcal{C}$ learns the inter-dependencies among the classes 
using a self-attention mechanism. 
For this, the input feature $X_{C_{in}} \in \mathbb{R}^{D_2\times T \times C}$ is first embedded using bottleneck convolutional layer (i.e. $1 \times 1$). 
After that, the output feature maps are rearranged into $\mathbb{R}^{D_2T \times C}$ and $\mathbb{R}^{C\times D_2T}$ followed by a matrix multiplication. The value of the resultant matrix is then normalized by a \texttt{softmax} activation. Now, the superimposed adjacency matrix $A_\mathcal{C}'$ can be formulated as:
\begin{equation}
\small
    A_\mathcal{C}'=A_\mathcal{C}+ softmax(W_{1}^{\top}X_{C_{in}}^{\top}W_{2}X_{C_{in}})
\end{equation}
where $X_{C_{in}}$ is the input of the C-GCN, and $W_1$ and $W_2$ are the weights of the bottleneck convolutions. 
Each value in this matrix can be seen as a soft edge between two vertices.
The learned graph is shared across different time-steps but unique for different layers and videos. This design choice can capture the inter-class dependencies in a video and makes C-GCN scalable across different temporal scales. 
Finally, we perform the graph convolutional operation with the formulation in~\cite{kipf2016semi}:
\begin{equation}
\small
    X_{C_{out}} = A'_\mathcal{C}X_{C_{in}}W_{3}
\end{equation}
{where $W_{3}\in \mathbb{R}^{D_2\times D_2}$ is the learnable weight matrix.} The operation with $A'_\mathcal{C}$ and with $W_3$ represents the message passing and vertex feature updating, respectively. Finally, $X_{C_{out}}$ is rearranged to $\mathbb{R}^{D_2 \times T \times C}$.

\subsubsection{CTM Block}
As shown in Fig.~\ref{Fig:overall}, there are $L$ blocks in CTM, each block is composed of a C-GCN and a TCN layer along with batch normalization and non-linear activations. 
To stabilize the training, two residual connections are added in each block. 

As mentioned earlier, TCN~\cite{lea2017temporal} aggregates the features across the temporal dimension while increasing the size of the temporal receptive field. 
In this work, we set a fixed kernel size $K$ for all the TCNs. 
%
%
Thanks to the hierarchical structure of CTM, C-GCN can focus on short-term action-dependencies in lower blocks and long-term action dependencies in higher blocks. 
The refined feature representation from the last block is fed into G-Classifier for the snippet-level classification.

\subsection{G-Classifier (G-Clf)}
\label{sec:g-classifier}
Finally, we introduce a graph-based G-Classifier to perform the final snippet-level classification.
In action detection, multiple actions could happen simultaneously; thus, prior knowledge of inter-dependencies among different action classes can benefit in making precise predictions. {Inspired from~\cite{chen2019multi} in image recognition, we introduce a GCN-based classifier in our task, which has different working mechanisms. 
In the previous work~\cite{chen2019multi}, the input of GCN is the word embedding of the label and the output is the classifier weights between the feature representation and the prediction.
In contrast to that, our input and output to the GCN are the video representation and the prediction scores respectively.  
As our input is directly the video representation, thus the action occurrence information within each snippet are captured, which benefits the information propagation between relevant classes through graph connections. 
Compared to the standard binary classifier~\cite{superevent}, G-Classifier has an additional message passing step between the potential co-occurring action pairs, thus improving the co-occurring action detection performance.} 
Different to C-GCN, G-Classifier models only the snippet-level features and focuses only on the actions that occur simultaneously.  

In practice, firstly, we compute the co-occurrence probabilities of all the action pairs in the training snippets. $M_{ij}$ indicates the concurring times for action class $C_i$ and $C_j$.
Then, the conditional probability matrix $P_{ij}=P(C_j|C_i)$ is given by:
\begin{equation}
\small
   P_{ij}=M_{ij}/N_i
\end{equation}
where $N_i$ indicates the occurrence times of $C_i$ in training set, and $P_{ij} \in \mathbb{R}^{C \times C}$ indicates the probability of class $C_j$ given that $C_i$ occurs at the same time.
%
In fine-grained action datasets, some rare co-occurrences may add noise for detecting other common actions, and the number of co-occurrences from training and test set may not be completely consistent. In this work, we perform a thresholding operation to binarize the conditional probability matrix to filter the noisy edges and make the classifier more robust to inconsistent action classes. If $P_{ij} \geq \theta$, $A_{S_{ij}}$ is assigned 1, otherwise 0, where $\theta$ is the threshold. 
%
%
The computed co-occurrence matrix $A_S$ is a binary correlation matrix which in turn defines the adjacency matrix of the graph for G-Classifier. 
The feature of a node is computed by the weighted sum of its own features and the adjacent nodes’ features. However, the binary correlation matrix may change the feature scale~\cite{kipf2016semi} and make the node feature over-smoothed~\cite{li2018deeper}. To alleviate this problem, we normalized the $A_S$ following the re-weighted scheme in~\cite{chen2019multi}.
Different to the learnable adjacency matrices in C-GCN, $A_S$ is fixed during training.
%
%
The formulation of this G-Classifier is given below:
\begin{equation}
\small
    S = \sigma(A_SX^{L}W_{S})
\end{equation}
where $S$ is the prediction score, $\sigma$ is the \texttt{sigmoid} activation. $X^{L}$ is the output feature from the last block of the Class-Temporal Module, and $W_{S}\in \mathbb{R}^{1\times D_2}$ are the learnable weights of the G-Classifier. 
To learn the parameters, we optimize the multi-label binary cross-entropy loss with the prediction results from the RTM and CTM. The total objective is formulate as:
\begin{equation}
\small
    \mathcal{L}_{total} = \mathcal{L}_{CTM} + \alpha \mathcal{L}_{RTM}
\end{equation}
where $\alpha$ is a weighting factor. Thus, by jointly optimizing both the entropy losses, the model learns the relevant action labels per segment along with learning the class-specific semantics across the Representation transform module.

\begin{table}[t!]\tabcolsep=14pt
\ttabbox{\caption{\small Comparison with the State-of-the-art on three densely labelled datasets. The results are given in per-frame mAP (\%). RGB
+OF indicates the late fusion performance. }}{
\scalebox{0.75}{
\label{tab:sota}
\begin{tabular}{l|c|c|c|c}
\hline 
Model           & Modality                   & Charades             & TSU  & MultiTHUMOS          \\\hline\hline
R-C3D~\cite{RC3d}            &  RGB                    & 12.7                 & 8.7  & -                    \\
I3D + TAN~\cite{dai2019tan}       &  RGB+OF                    & 17.6                 & -    & 33.3                 \\
I3D + Superevent~\cite{superevent}&  RGB                    & 18.6                 & 17.2 & 36.4                 \\
I3D + TGM~\cite{TGM1}       &  RGB                       &20.6                      & 26.7  &37.2                     \\
I3D + TGM~\cite{TGM1}       &  RGB+OF                    & 21.5                 & -    & 44.3                 \\\hline
I3D + TGM + Superevent~\cite{TGM1}       &  RGB+OF                    & 22.3                 & -    & 46.4                 \\\hline
I3D + MLAD~\cite{MLAD}      &  RGB                       & 18.4                 & -    &  42.2                    \\
I3D + MLAD~\cite{MLAD}      &  RGB+OF                    & 22.9                 & -    &   49.6                   \\\hline
I3D + CTRN      &  RGB      & 25.3                       & \textbf{33.5}                 & 44.0                      \\
I3D + CTRN      &  RGB+OF   & \textbf{27.8}                       & -                    & \textbf{51.2}                     \\\hline
\end{tabular}}}
\end{table}

\begin{figure}[t!]
\begin{floatrow} 
  \begin{floatrow}    
\ffigbox{%
  \includegraphics[width=0.9\linewidth]{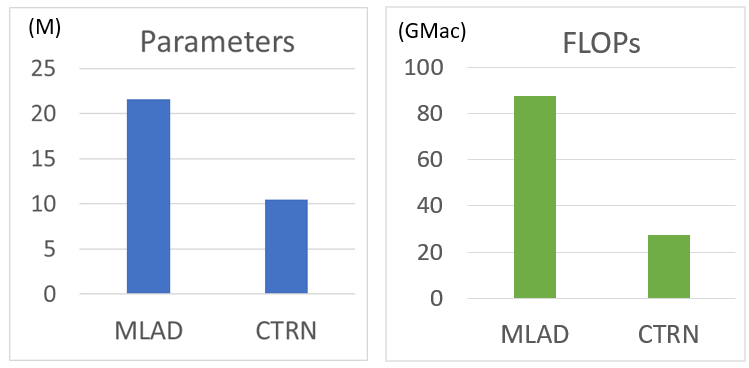}
  \label{Computation_cost}
}{
  \caption[0.5]{Computation efficiency.}
}
\end{floatrow}

\ffigbox{%
  \includegraphics[width=.85\linewidth]{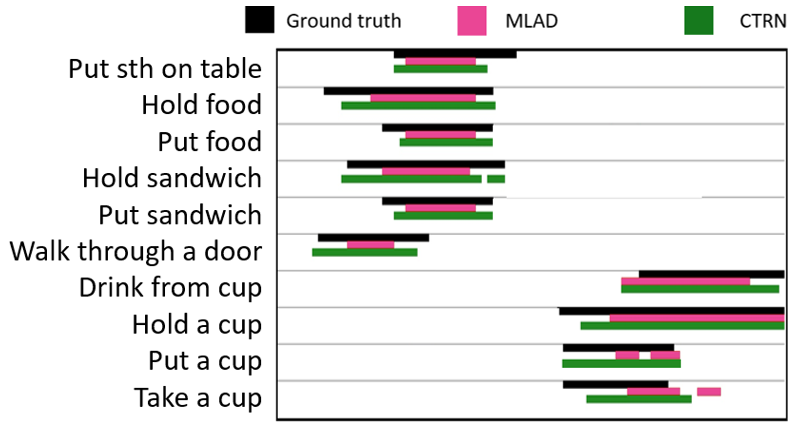}
  \label{Detection_Performance}
}{%
  \caption[]{Model prediction visualization.}%
}
\end{floatrow}
\end{figure}

\section{Experiment}
\subsection{Experimental Settings}
\textbf{Datasets.}
We evaluate our method on three densely labelled action detection datasets: Charades~\cite{sigurdsson2017asynchronous}, TSU~\cite{dai2020toyota} and MultiTHUMOS~\cite{multi-thumos}. 
These datasets contain videos of different types: (1) sports and daily living videos, (2) short and long videos. 
We follow the original settings of these datasets for action detection. All these datasets are evaluated by the per-frame mAP.\\
%
\textbf{Implementation.}
To have a fair comparison with previous works~\cite{TGM1,MLAD,superevent}, our network is built on top of I3D, where $D_1$ is 1024 and $D_2$ is 64. Dropout probability is 0.3. For CTM, we choose a 5-block ($L$) structure. For C-GCN, the adjacency matrix is initialized by 1 and normalized by columns. 
In TCN, the kernel size $K$ is 9 and padding rate is 4. For G-Classifier, $\theta$ is set to 0.05. While learning the parameters, the weighting factor $\alpha$ is 1.2 and the random seed is fixed.   
We use Adam optimizer~\cite{adam_optimizer} with an initial learning rate of 0.001, and we scale it by a factor of 0.3 with a patience of 10 epochs. The network is trained on a 4-GPU machine for 300 epochs. For two-stream network, a mean pooling is performed between the prediction logits of the RGB and Flow streams.

\begin{table}[t!]\tabcolsep=9pt
\caption{Evaluation on the Charades dataset using the action-conditional metric~\cite{MLAD}. $P_{AC}$ - Action-Conditional Precision, $R_{AC}$ - Action-Conditional Recall, $F1_{AC}$ - Action-Conditional F1-Score, $mAP_{AC}$ - Action-Conditional Mean Average Precision. $\tau$ indicates the temporal window size. $\tau = 0$ corresponds to the actions occuring at the same time.}
\scalebox{0.68}{
\label{tab-cooccuring}
\begin{tabular}{l|cccc|cccc|cccc}
\hline
     & \multicolumn{4}{|c|}{$\tau$ = 0} & \multicolumn{4}{|c|}{$\tau$ = 20} & \multicolumn{4}{|c}{$\tau$ = 40} \\\hline
     & $P_{AC}$   & $R_{AC}$   & $F1_{AC}$   & $mAP_{AC}$  & $P_{AC}$   & $R_{AC}$   & $F1_{AC}$   & $mAP_{AC}$   & $P_{AC}$   & $R_{AC}$   & $F1_{AC}$   & $mAP_{AC}$   \\\hline
I3D  &14.3     &1.3     &2.1      &15.2      &12.7     &1.9     &2.9      &21.4       &14.9     &2.0     &3.1      &20.3       \\
CF   &10.3     &1.0     &1.6      &15.8      &9.0     &1.5     &2.2      &22.2       &10.7     &1.6     &2.4      &21.0       \\
MLAD~\cite{MLAD} &19.3     &7.2     &8.9      &28.9      &18.9     &8.9     &10.5      &35.7       &19.6     &9.0     &10.8      &34.8       \\\hline
CTRN &\textbf{23.9}     &\textbf{8.0}     &\textbf{11.9}     &\textbf{29.7}      &\textbf{21.7}     &\textbf{9.1}     &\textbf{12.9}      &\textbf{36.8}       &\textbf{23.0}     &\textbf{9.3}     &\textbf{13.2}      &\textbf{35.5}      \\\hline
\end{tabular}}
\end{table}

\subsection{Comparison with State-of-the-Art Methods}
The proposed CTRN is compared with previous state-of-the-art methods 
on the Charades, TSU and MultiTHUMOS datasets in Table~\ref{tab:sota}.
Our proposed method outperforms current state-of-the-art methods on all three datasets. For example, +6.9\% (relatively +37.5\%) w.r.t. MLAD~\cite{MLAD} on Charades while using only RGB. We then show the ability of CTRN capturing action co-occurrence, we evaluate with the action-conditional metric~\cite{MLAD} in Table~\ref{tab-cooccuring}. Compared with state-of-the-art methods, our method achieves higher performance on all action-conditional metrics showing that CTRN effectively models action dependencies both within a time-step (i.e. co-occurring action, $\tau$ = 0) and throughout time ($\tau$ > 0). 

To confirm the advancement of our method, we present further comparisons with MLAD. 
Firstly, we compare the model efficiency and complexity in Fig.~{\color{red}3}. MLAD is about 2 times larger in parameters and 3.5 times larger in FLOPs than CTRN while processing the same batch of videos. Hence, our method is more lightweight and computationally efficient than MLAD. 
Secondly, we visualize the detection results of an example video in Fig.~{\color{red}4} for both MLAD and CTRN. 
Qualitatively, we find that CTRN can detect the actions more precisely than MLAD. 
%

%

\begin{figure}
\begin{floatrow}
\capbtabbox{%
\scalebox{0.65}{
\begin{tabular}{ccccc}
\hline
\multicolumn{4}{c}{CTRN Components}                                                                             & \multicolumn{1}{c}{Charades} \\
RTM                  & \multicolumn{1}{c}{C-GCN} & \multicolumn{1}{c}{TCN} & \multicolumn{1}{c}{G-Classifier} & \multicolumn{1}{c}{Per-frame mAP}      \\\hline
\multicolumn{1}{c}{$\times$} & $\times$& $\times$& $\times$       & 15.6   \\\hline
\checkmark                    & $\times$& $\times$& $\times$       & 16.1    \\\hline
$\checkmark$                    &$\checkmark$     & $\times$& $\times$ & 19.9    \\
$\checkmark$                    & $\times$&$\checkmark$     & $\times$  &21.4    \\
$\checkmark$                    &$\checkmark$     &$\checkmark$     & $\times$  & 24.7    \\\hline
$\checkmark$                    & $\times$& $\times$&$\checkmark$  & 18.4    \\
$\checkmark$                    &$\checkmark$     &$\checkmark$     &$\checkmark$   & \textbf{25.3}  \\\hline
\end{tabular}
}
\vspace{-2mm}
}{%
  \caption{Ablation study on Charades dataset using only RGB. }%
  \label{table:ablation}
}

\ffigbox{%
  \includegraphics[width=0.5\linewidth]{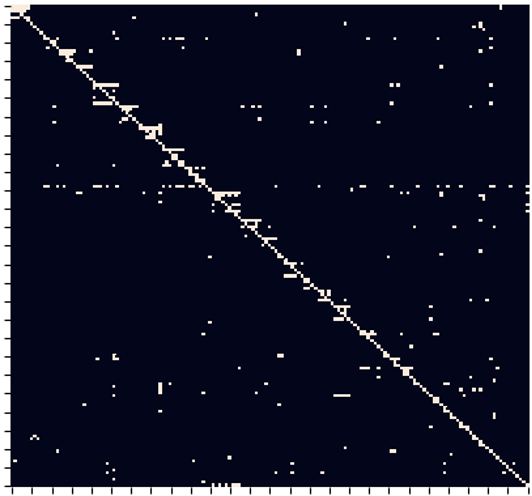}
  \vspace{-3mm}
}{%
  \caption{The adjacency matrix of the G-Classifier $A_S$.}
}
\end{floatrow}
\end{figure}

\begin{figure}[t]
\begin{floatrow}
\capbtabbox{\tabcolsep=16pt
\scalebox{0.68}{
\begin{tabular}{cc|ccc}
\hline
\multicolumn{2}{c}{CTM Component} & \multicolumn{3}{|c}{\#Block} \\
C-GCN             & TCN             & 1        & 3       & 5       \\\hline
$\checkmark$      &                 & 17.9     &19.3     & 20.0         \\
                  & $\checkmark$    & 18.7     &20.1     & 21.9         \\
$\checkmark$      & $\checkmark$    & 19.8     &23.3     & 25.3    \\\hline   
\end{tabular}}
}
{%
  \caption{Analysis between C-GCN and TCN. }\label{Tab:TCN-C-GCN}
}

\capbtabbox{\tabcolsep=15pt
\scalebox{0.83}{
\begin{tabular}{lcc}
\hline
               & \multicolumn{2}{c}{Backbone} \\ 
               & RTM          & CTRN          \\\hline
w/o Threshold  & 17.8         & 24.9               \\
with Threshold & 18.4         & 25.3              \\ \hline
\end{tabular}
}
}{%
  \caption{Ablation study on threshold. }\label{table:threshold}
}
\end{floatrow}
\end{figure}

\subsection{Ablation studies}
Firstly, in Table~\ref{table:ablation}, we study the complementation of the components in the proposed network on the Charades dataset. 
We first discuss how RTM leads to a better feature representation of the input spatio-temporal feature map from I3D. RTM is an essential pre-step before class-temporal modelling. Thanks to RTM that filters the class-specific feature, the model can slightly improve the detection performance (+0.5\%). We then explore how the different components in CTM affects the action detection performance. We find that both C-GCN and TCN improve the performance w.r.t. a model with only RTM (+23.6, and 32.9\% relatively). The action detection performance is further improved by the combination of both C-GCN and TCN, thus reflecting the complementary nature of both the operations. Finally, we study the performance with/without G-Classifier. With the proposed classifier, RTM and RTM+CTM further improve the action detection performance by +2.3\% and +0.6\% respectively. Note that for the baseline without G-Classifier, similar to the previous work~\cite{superevent}, we utilize a $1\times 1$ convolution as the classifier. %
These results show that the different components of CTRN contribute to the overall performance of our network.

{Secondly, we analyse if G-Classifier can better detect the co-occurring actions. We compared our method with the standard binary classifier~\cite{superevent}, Chen et al.~\cite{chen2019multi} by computing the mAP with the snippets containing more than one action. 
All three classifiers are build upon the same backbone (CTRN). We find that the standard binary classifier, Chen et al. and G-Classifier achieve 24.1\%, 25.3\% and 27.6\% respectively on Charades, reflecting that G-Classifier can better detect the co-occurring actions.}

To further explore the relationship between inter-classes and temporal modules, we perform ablations with C-GCN and TCN having different numbers of blocks. In Table~\ref{Tab:TCN-C-GCN}, we find that while having more blocks: C-GCN improves marginally since it captures action relations in a single temporal scale; TCN improves the performance with more blocks due to increased temporal receptive field; C-GCN+TCN achieves the best performance than TCN since C-GCN can capture action relations in more temporal scales when combined with TCN (see the improvements for 5 blocks). 

Finally, we explore how hard thresholding affects the performance of G-Classifier. In Table~\ref{table:threshold}, we find that G-Classifier performs better with the threshold since hard threshold filters the noisy edges.


\begin{figure}[t]
\vspace{1mm}
\includegraphics[width=12.5cm]{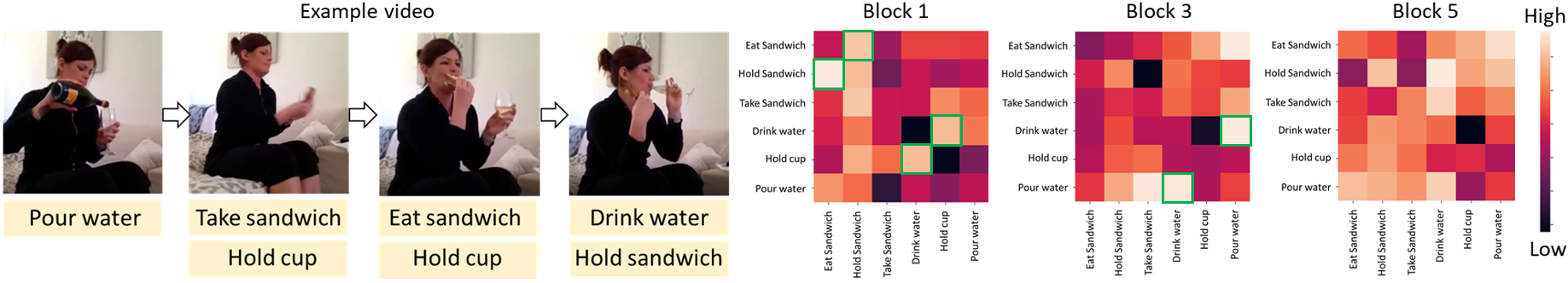}
\caption{Visualization of the learned C-GCN adjacency matrix $A'_\mathcal{C}$ for different layers. Here, we visualize the $1^{st}$, $3^{rd}$ and $5^{th}$ block's adjacency matrices. For simplicity, we provide only the relevant action classes in the example video. }
\label{Fig:cgcn_visualization}
\end{figure}

\subsection{Qualitative Analysis}
In Fig.~{\color{red}5}, we show the adjacency matrix of G-Classifier in Charades (157 classes), which provides the information of all the co-occurring action pairs with high probabilities.
For example, \textit{holding a vacuum} \& \textit{tiding something on the floor} and \textit{fixing hair} \& \textit{watching in a mirror} are the actions always occur at the same time.
Prior access to such privilege knowledge is crucial for detecting the co-occurring actions in the densely labelled videos.

In CTM, TCN is used to aggregate the temporal information which enables C-GCN to explore action relations at different temporal scales. 
{To validate the usage of these layers, in Fig.~\ref{Fig:cgcn_visualization}, we visualize the learned adjacency matrix of C-GCN from three different blocks.} 
We find that in Block 1, C-GCN focuses on capturing the contextual information pertaining to locally related action classes. For example, \textit{eat sandwich} \& \textit{hold sandwich} and \textit{drink water} \& \textit{hold cup} are always occurring closely in the video. 
Then we find that, Block 3 has increased the temporal receptive field, thus, C-GCN can capture the long-term dependencies between distant action classes. For example, \textit{Pour water} and \textit{Drink water}. 
{Finally, Block 5 possess the largest receptive field where each local snippet feature contains the whole video information. 
Therefore, C-GCN in this block models all the potential action relations in the video, resulting in many activated links in the adjacency matrix. }

\section{Conclusion}
{In this work, we propose a novel class-temporal relation network for action detection.
This network can handle both action class relations at the same time-step, and across different time-steps by using its three key components, namely Representation Transform Module, Class-Temporal Module, and G-Classifier.
The network is evaluated on three challenging densely labelled datasets and achieves state-of-the-art performance on all of them. Furthermore, this network has less computation cost and parameters than the representative baseline method. 
For future perspectives, we will investigate combining C-GCN and TCN into a single layer.}

\subsection*{Acknowledgement}
This work has been supported by the French government, through the 3IA Cote d’Azur Investments in the Future project managed by the National Research Agency (ANR) with the reference number ANR-19-P3IA-0002. The authors are also grateful to the OPAL infrastructure from Université Côte d’Azur for providing resources and support.
\bibliography{egbib}
\end{document}